# Automatic Diagnosis of Myocarditis Disease in Cardiac MRI Modality using Deep Transformers and Explainable Artificial Intelligence

Mahboobeh Jafari[1,2,*], Afshin Shoeibi[1,2], Navid Ghassemi[1], Jonathan Heras[3], Sai Ho Ling[4], Amin Beheshti[5], Yu-Dong Zhang[6], Shui-Hua Wang[6], Roohallah Alizadehsani[7], Juan M. Gorriz[2,8], U. Rajendra Acharya[9], Hamid Alinejad Rokny[10,11,12]

[1] Internship in BioMedical Machine Learning Lab, The Graduate School of Biomedical Engineering, UNSW Sydney, Sydney, NSW, 2052, Australia.

[2] Data Science and Computational Intelligence Institute, University of Granada, Spain.

[3] Department of Mathematics and Computer Science, University of La Rioja, La Rioja, Spain.

[4] Faculty of Engineering and IT, University of Technology Sydney (UTS), Australia.

[5] Data Analytics Lab, Department of Computing, Macquarie University, Sydney, NSW 2109, Australia.

[6] School of Computing and Mathematical Sciences, University of Leicester, Leicester, UK.

[7] Intelligent for Systems Research and Innovation (IISRI), Deakin University, Victoria 3217, Australia.

[8] Department of Psychiatry, University of Cambridge, UK.

[9] School of Mathematics, Physics and Computing, University of Southern Queensland, Springfield, Australia.

[10] BioMedical Machine Learning Lab, The Graduate School of Biomedical Engineering, UNSW Sydney, Sydney, NSW, 2052, Australia.

[11] UNSW Data Science Hub, The University of New South Wales, Sydney, NSW, 2052, Australia.

[12] Health Data Analytics Program, AI-enabled Processes (AIP) Research Centre, Macquarie University, Sydney, 2109, Australia.

* Corresponding author: Afshin Shoeibi (Afshin.shoeibi@gmail.com)

**Abstract**

Myocarditis is a significant cardiovascular disease (CVD) that poses a threat to the health of many individuals by causing damage to the myocardium. The occurrence of microbes and viruses, including the likes of HIV, plays a crucial role in the development of myocarditis disease (MCD). The images produced during cardiac magnetic resonance imaging (CMRI) scans are low contrast, which can make it challenging to diagnose cardiovascular diseases. In other hand, checking numerous CMRI slices for each CVD patient can be a challenging task for medical doctors. To overcome the existing challenges, researchers have suggested the use of artificial intelligence (AI)-based computer-aided diagnosis systems (CADS). The presented paper outlines a CADS for the detection of MCD from CMR images, utilizing deep learning (DL) methods. The proposed CADS consists of several steps, including dataset, preprocessing, feature extraction, classification, and post-processing. First, the Z-Alizadeh dataset was selected for the experiments. Subsequently, the CMR images underwent various preprocessing steps, including denoising, resizing, as well as data augmentation (DA) via CutMix and MixUp techniques. In the following, the most current deep pre-trained and transformer models are used for feature extraction and classification on the CMR images. The findings of our study reveal that transformer models exhibit superior performance in detecting MCD as opposed to pre-trained architectures. In terms of DL architectures, the Turbulence Neural Transformer (TNT) model exhibited impressive accuracy, reaching 99.73% utilizing a 10-fold cross-validation approach. Additionally, to pinpoint areas of suspicion for MCD in CMRI images, the Explainable-based Grad Cam method was employed.

## 1. Introduction

The prevalence of cardiovascular diseases (CVDs) is now a major global public health issue [1]. According to the World Health Organization (WHO), CVDs currently rank among the leading causes of mortality globally [2-3]. The heart is responsible for the circulation of blood, carrying oxygen and nutrients throughout the body, and expelling carbon dioxide [4]. The most common CVDs include coronary artery disease (CAD) [5], arrhythmia [6], cardiomyopathy [7], heart failure [8], congenital heart disease [9], mitral regurgitation [10], angina [11], and myocarditis disease (MCD) [12]. Symptoms of CVDs usually manifest in chest pain, arm and left shoulder pain, shortness of breath, nausea and fatigue, cold sweats, headache, and dizziness [13-15]. Risk factors for CVDs include hypertension, smoking, high cholesterol levels, diabetes, obesity, family history, and advanced age [16-18]. Clinical studies have demonstrated that minimizing these risk factors significantly reduces the likelihood of CVDs occurrence and ensures good health [1-4].

In recent years, the incidence of MCD has increased worldwide [19-22]. MCD is caused by inflammation of the myocardium, which impairs the heart's ability to pump blood throughout the body, leading to significant health issues such as arrhythmia, chest pain, and dyspnea in patients [19-20]. In some cases, myocardial dysfunction can also result in blood clots, heart attack, stroke, heart injury, heart failure, or even death [21-22]. Clinical studies have revealed that SARS-CoV-2, adenovirus, and HIV are the primary causes of MCD [19-20]. While MCD may be asymptomatic in some cases, it often causes chest pain, heart failure, fever, palpitations, fatigue, and even sudden death [21]. Medical professionals use various screening methods to diagnose CVDs, including ECG, Echo, cardiac exercise testing, CT, CMRI, and Holter monitoring [23-28]. Among these methods, CMRI imaging is considered one of the most effective non-invasive methods for diagnosing CVDs, including MCD. Specialists use CMRI images to analyze different heart regions, such as ventricular wall thickness and left ventricular end-systolic volume [29-31]. CMRI imaging provides early diagnosis of cardiac dysfunction, enabling rapid and reliable diagnosis of CVDs, including MCD [29-31]. Despite the advantages of CMRI, analyzing medical images is a challenging task for doctors. For accurate diagnosis of MCD multi-slice CMRI images must be acquired for each patient, which is typically a difficult task even for experienced doctors [20-21]. Additionally, CMRI images may have low resolution and low contrast, making it challenging to diagnose MCD [21]. Furthermore, CMRI images may show various artifacts that complicate the diagnosis of MCDs by specialists.

To address these challenges, researchers have used Artificial Intelligence (AI) techniques to diagnose MCD from CMR images [29-31]. AI techniques are generally categorized into ML and DL methods [95-96]. In references [97-98], research on the diagnosis of CVDs from CMRI data using ML techniques is reported. In this research, ML-based CADS involves dataset, pre-processing, feature extraction, feature selection and classification stages [97-98]. The researchers of these papers have demonstrated that the feature extraction is the most important part in ML-based CADS for the diagnosis of CVDs [97-98]. As such, subsequent to the phases of data registration and preprocessing, researchers attempt to improve the diagnosis of CVDs by combining different feature extraction methods. However, combining multiple features to achieve high diagnosis accuracy requires extensive knowledge of researchers in the field of ML [97]. To address this challenge, DL techniques have recently been proposed for detecting or predicting CVDs such as MCD [99-100]. Unlike ML methods, DL techniques employ feature engineering in an unsupervised manner. References [99-100] have reviewed various papers on the diagnosis of CVDs from CMR images using DL techniques. These studies demonstrate that researchers using DL techniques have achieved significant results in the classification and segmentation of CMR images for the diagnosis of CVDs, including MCD [99-100].

In recent years, studies on the diagnosis of MCD from CMRI images using DL techniques have been published [19-21]. Sharifrazi et al. [19] proposed the CNN-KCL model for MCD detection based on CMRI images, and experiments were performed on the Z-Alizadeh dataset. The study aimed to combine a 2D-CNN model with the k-means clustering method. Their findings showed an accuracy of 97.41%. In another study by Shoeibi et al. [20], the cycle-GAN method was utilized with various pre-trained models to diagnose MCD. The Z-Alizadeh dataset was also used in this study to implement the proposed model. The cycle-GAN architecture was employed in the preprocessing step to develop synthetic CMRI images, which were then applied to various pre-trained models. Among them, the EfficientNet V2 method achieved an accuracy of 99.33%. Moravvej et al. [21] introduced deep reinforcement learning (RL) for MCD detection using CMRI images and presented an RLMD-PA method to diagnose myocarditis. Lastly, several optimization methods were evaluated to enhance the accuracy and efficiency of MCD diagnosis.

In this study, we propose a deep learning-based computer-aided diagnosis system (CADS) for MCD diagnosis from CMRI images. The presented paper introduces various novelties in the pre-processing, classification and post-processing sections. In the first novelty, we utilized CutMix and MixUp techniques in the pre-processing section [32-33]. Notably, this method has not been previously employed in other studies for to generate artificial CMR images to improve the accuracy of CVDs diagnosis [97-100]. Additionally, transformers architectures are a new class of DL techniques that have shown high performance in various medical imaging and signal processing applications [101-102]. Compared to other DL architectures, transformers techniques show high performance on low input images [101-102]. As a second novelty, we have simulated and compared the latest transformers architectures. Additionally, we compared these networks with the most recent pretrained models to demonstrate the efficiency of our proposed method. We thoroughly reviewed various papers in the field of CVDs diagnosis, including myocarditis, and found that transformers methods have not been used by researchers [97-100]. The results demonstrated that we achieved the highest accuracy in diagnosing MCD from CMR images by using the combination of CutMix and MixUp methods and some transformers models. Another novelty is the use of an explainable artificial intelligence technique as Grad-Cam, which is important for clinical diagnoses [103]. XAI is a rapidly growing field in medicine, and current research is focused on utilizing these methods to diagnose various diseases from medical images [104-105]. In this study, we employed the Grad-Cam technique as an XAI method, which provides a detailed understanding of the performance of DL architectures used in the diagnosis of myocarditis from CMR images. These assist specialist physicians in accurately diagnosing myocarditis from CMR images.

Some of the authors of this paper have previously published a comprehensive review paper on the diagnosis of cardiac diseases from CMR images using DL techniques [106]. This review paper highlights that a limited number of researchers have employed XAI techniques in CVDs diagnosis. Moreover, the reference [106] and other review papers in this field [97-100] indicate that the proposed method, including CutMix and MixUp methods, transformer models and XAI, has not been previously employed for the diagnosis of myocarditis from CMR images. The rest of the paper is categorized as follow: Section 2 describes the proposed CADS in detail, including the dataset, preprocessing, and DL model. Section 3 represents the evaluation parameters of the proposed method. The experiment results with details are presented in Section 4. Section 5 is about post-processing. In section 6, we discussed about limitation of study. The discussion section of the paper is presented in section 7. Finally, the conclusion and future directions are provided in Section 8.

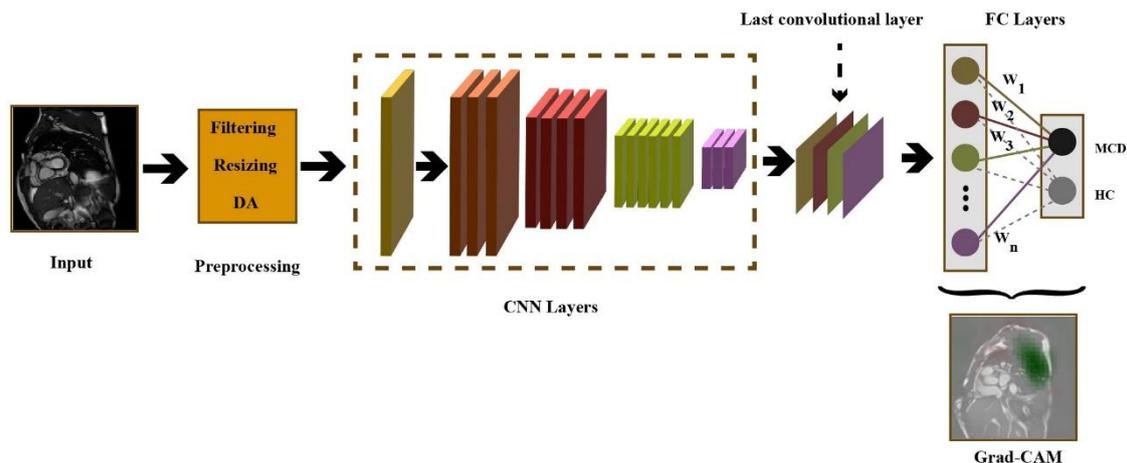

Fig. 1. Block diagram of the proposed method for diagnosis of MCD

## 2. Material and Methods

This section proposes a CADS to diagnose MCD using CMRI images. Figure (1) shows the block diagram of the proposed method which includes dataset, preprocessing, DL model, and post-processing. First, a DL model is implemented using the Z-Alizadeh dataset [19]. The Z-Alizadeh dataset consists of 12,000 CMRI images of normal and MCD patients admitted to Shahid Rajaee Hospital, Tehran. In the pre-processing step, denoising, resizing, and a new DA method to generate synthetic CMRI images were performed. The CMRI images were denoised in this section and then resized to 224*224. In the following, synthetic CMRI data was generated using a new DA model based on the CutMix [32] and MixUp [33] approaches. The proposed DA method is used for the first time in MCD detection and is the first novelty in this study. In the third step, state-of-the-art pre-trained models and transformers were used for feature extraction and classification of CMR images. Pre-trained models included EfficientNet B3 [35], EfficientNet V2 [36], HrNet [37], Inception [38], ResNetrs50 [39], ResNest50d [40], and ResNet 50d [41]. In addition, the transformer models also included Beit [42], Cait [43], Coat [44], Deit [45], Pit [46], Swin [47], TNT [48], Visformer [49], and ViT [50]. The transformer model for diagnosing MCD is another novelty of this work. Finally, the Grad-Cam technique, an explainable AI technique, was used to visualize the suspicious regions of MCD in CMRI images [34].

### 2.1. Z-Alizadeh Dataset

The Z-Alizadeh Sani myocarditis dataset was collected between September 2018 and September 2019 at the CMR department of OMID hospital in Tehran, Iran. The soundness of the data gathering process has been confirmed by the local ethical committee of OMID hospital. To perform CMR examination, a 1.5-T system (MAGNETOM Aera Siemens, Erlangen Germany) was used. Dedicated body coils were used to scan each patient in the standard supine position. The CMR protocols that have been complied with are listed below:

* CINE-segmented images and pre-contrast T2-weighted (trim) images were performed in short and long axes views.
* The pre-contrast T1-weighted relative images were acquired in axial views of the myocardium.
* After injection of Gadolinium ((DOTAREM 0/1 mmol/kg), the T1-weighted relative sequence was repeated. After 10-15 minutes, sequences of Late Gadolinium Enhancements (LGE- high-resolution PSIR) in short and long axes views were carried out.

The total number of images is 10425. The number of images representing HC and MCD patients were 5040 and 5385, respectively. Figure (2) shows typical CMRI images obtained from the Z-Alizadeh dataset for healthy control (HC) and MCD patients.

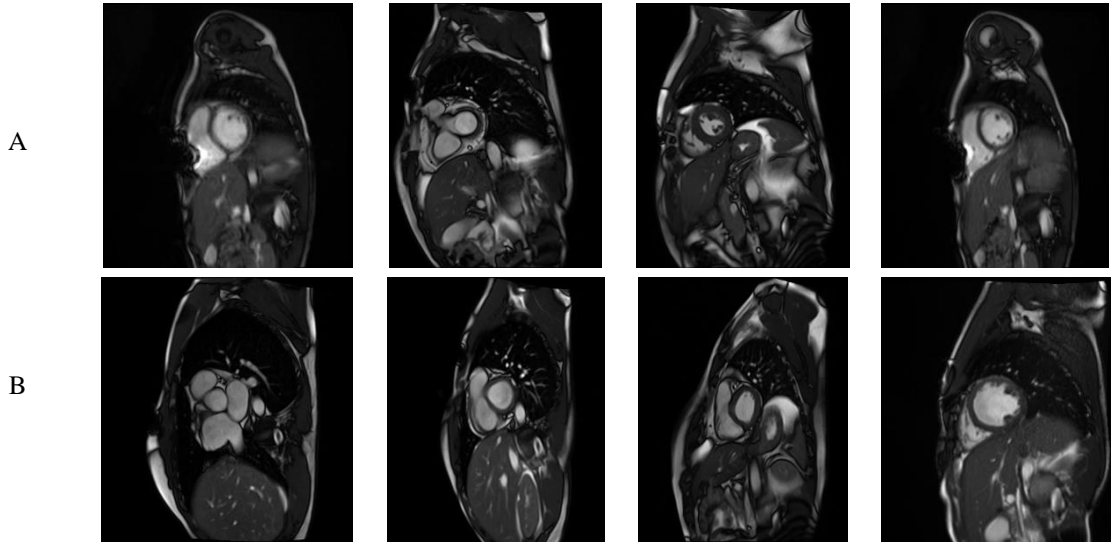

Fig 2. Sample CMR images dataset: a) Normal, and B) Abnormal.

## 2.2. Preprocessing

This section describes preprocessing steps used for CMRI images, involving denoising, image resizing, and DA. Preprocessing is a crucial step in the analysis of medical images, particularly MRI data. In this work, pre-processing steps entail noise removal, image size reduction and DA. CMR images contain various artifacts during registration, posing a challenge for specialist physicians to accurately diagnose myocarditis. Additionally, some slices of CMR images may have low contrast for certain subjects. To diagnose myocarditis, specialist doctors record CMR images of different regions of the heart, resulting in a dataset of images with varying sizes. To address this issue, the next preprocessing step involves resizing all images to 224x224 dimensions. This will allow uniform image sizes to be applied to the DL networks. The selection of 224x224 image size was made through a trial-and-error process in this preprocessing stage, ensuring that reducing the size of CMR images does not lead to a reduction in evaluation parameters. It was observed that reducing image sizes results in a decrease in hardware resource usage. However, transformer models have a comparatively long training time when compared to pretrained models. By reducing image size to 224x224, training time for these networks has significantly decreased while simultaneously utilizing fewer hardware resources. In the continuation of the pre-processing section, CutMix and MixUp techniques have been employed for data augmentation (DA) [32-33]. This preprocessing approach represents the first novelty of this paper. In references [97-100], review papers in the field of CVDs diagnosis from CMR images using AI techniques are reported. It can be seen that this method has not been utilized for DA in similar research studies, and most researchers have usually used GAN techniques for this task. In the following, the details of the CutMix and MixUp methods are provided for the purpose of DA.

## 2.3. Data Augmentation

Data augmentation is a technique used in DL to increase the amount of training data by creating new samples from existing ones [107-108]. This is done by applying a set of transformations to the original data, creating variations that the model can learn from. DA is particularly useful when the amount of training data is limited, as it can help prevent overfitting and improve model performance [51]. In medical applications, researchers frequently encounter the challenge of limited access to medical images containing a large number of subjects. To address this challenge, researchers have introduced various methodologies, of which DA techniques are one of the most important [107-108]. References [106-109] indicate that researchers have employed DA techniques such as different generative adversarial network (GAN) models to diagnosis types of CVDs from CMR images. The results of these studies demonstrate that the utilization of DA methods has led to an improvement in the accuracy and efficiency of CVD diagnosis from CMR images. Although GAN techniques have proven effective in enhancing the efficiency and accuracy of medical diagnosis, the training of these methods is relatively complex [107-108]. In this work, CutMix and MixUp methods [32-33] have been used for DA.

CutMix and MixUp methods [32-33] are one of the latest DA techniques that have recently been successfully employed in various medical applications. As demonstrated in reference [109], researchers have utilized CutMix and MixUp techniques in the diagnosis of brain tumors. In this study, DA was initially conducted using this methodology, followed by utilizing a DL architecture to segment brain MR images. In another research, CutMix and MixUp methods were employed for brain lesion segmentation, yielding satisfactory outcomes [110]. Rana et al. [111] employed CutMix and MixUp methods for blood cell image classification and achieved acceptable results.

CutMix replaces a part of one image with a part of a different image, while MixUp generates an example by mixing two images and their corresponding labels [32-33]. Namely, given two random images from the dataset, $x_i$ and $x_j$, and their corresponding one-hot labels, $y_i$ and $y_j$, CutMix constructs a new training example $(x, y)$ by using the following formulas:

$$x = M \times x_i + (1 - M) \times x_j$$

$$y = \lambda \times y_i + (1 - \lambda) \times y_j$$

where $\lambda$ are values between [0, 1] range and are sampled from the Beta distribution; M denotes a binary mask indicating where to drop out and fill in from two images. To sample the binary mask M, CutMix samples the bounding box coordinates $B = (r_x, r_y, r_w, r_h)$ indicating the cropping regions on $x_i$ and $x_j$. The region B in $x_i$ is removed and filled in with the patch cropped from B of $x_j$. Following the approach of [32], we sampled rectangular mask M whose aspect ratio is proportional to the original image. The box coordinates are uniformly sampled according to [32]:

$$r_x \; Unif(0, W)$$

$$r_y \; Unif(0, H)$$

$$r_w = W \times \sqrt{1 - \lambda}$$

$$r_h = H \times \sqrt{1 - \lambda}$$

where W and H are the weight and height of the original image [32], respectively. Figure 3 shows the CutMix method applied to CMRI images.

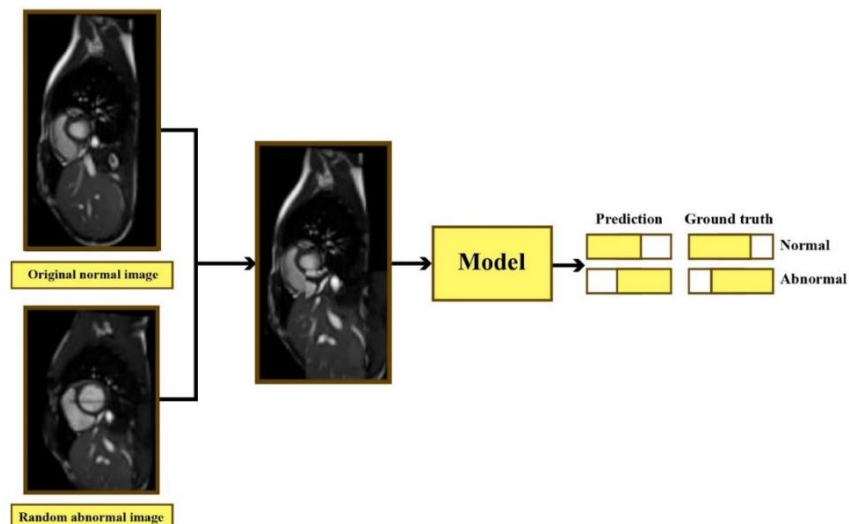

Fig. 3. Block diagram of CutMix method for DA from CMRI images.

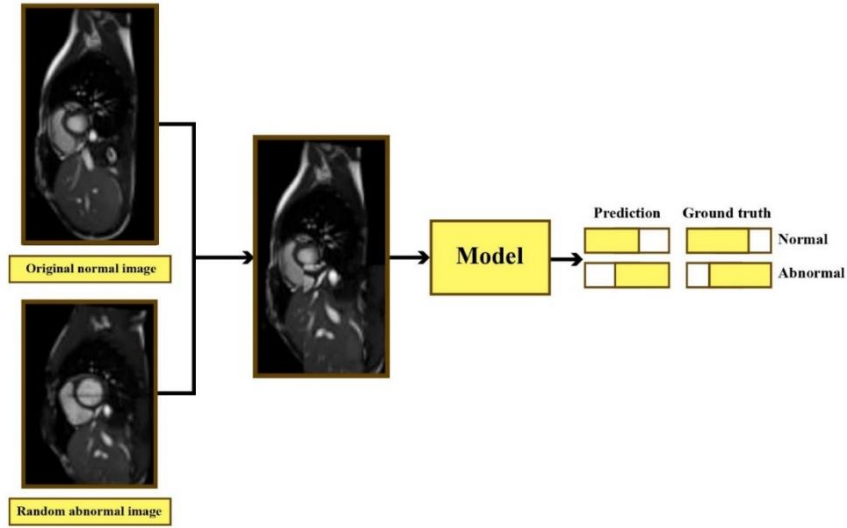
Fig. 4. Block diagram of MixUp method for DA from CMRI images.

MixUp [33] also uses information from two images, but instead of implanting one portion of an image inside another, MixUp produces an elementwise convex combination of two images. Namely, given two random images from the dataset, $x_i$ and $x_j$, and their corresponding one-hot labels, $y_i$, and $y_j$, MixUp constructs a new training example $(x, y)$ by using the following formulas [33]:

$$x = \lambda \times x_i + (1 - \lambda) \times x_j$$
$$y = \lambda \times y_i + (1 - \lambda) \times y_j$$

where $\lambda$ are values between [0, 1] range and are sampled from the Beta distribution [33]. Figure 4 shows the MixUp method applied to CMRI images.

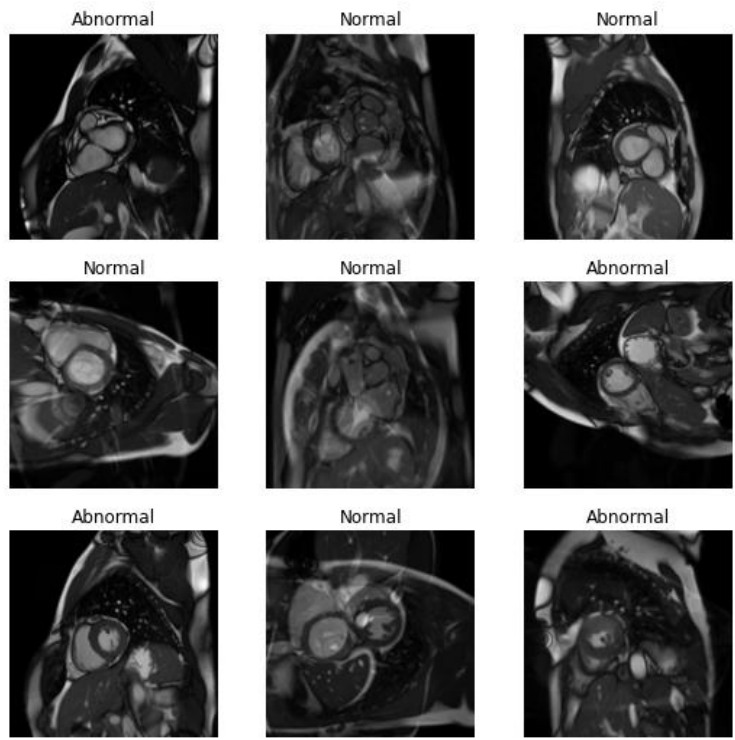
Fig. 5. Some CMRI images generated using the proposed DA method.

CutMix [32] and MixUp [33] prevent overfitting during training and improve the model's ability to generalize to out-of-distribution examples. In this study, we have used these techniques using the callbacks provided by the FastAI library [52]. Figure (5) shows a sample of the synthetic CMRI images generated using these techniques for both HC control and MCD classes.

### 2.4. Deep Learning Models

In recent years, extensive research has been conducted in the field of diagnosing CVDs from CMR images, employing various ML and DL techniques [97-100]. In papers on the diagnosis of CVDs using ML techniques, researchers have utilized handcrafted feature extraction methods in the diagnosis of CVDs from CMRI data and have reached satisfactory results [97-98]. In references [97-98], A number of review papers on the diagnosis of CVDs from CMR images using ML techniques are mentioned. It is evident that feature extraction and selection methods are the most important components of ML-based CADS for CVDs diagnosis [97]. Based on the literature [97-98], the most prominent handcrafted feature extraction techniques for CMR images encompass Gabor [112], Haralick [113], and local binary patterns (LBPs) [114]. In ML-based CADS, feature extraction and selection methods are typically chosen through trial and error, with the aim of enhancing the accuracy of CVDs diagnosis from CMR data. This process is very time-consuming and requires high knowledge of researchers in the field of ML and especially feature extraction techniques [106]. Moreover, ML techniques usually tend to underperform when confronted with a large volume of input data, while many images of patients with various CVDs are recorded for accurate diagnosis [106]. These challenges have presented significant obstacles for researchers in their pursuit of developing practical software solutions for the diagnosis of a range of CVDs, including myocarditis, utilizing ML techniques [106].

In order to address the challenges of ML methods, DL networks have been introduced and rapidly applied in various medical applications, including the diagnosis of CVDs from CMRI data [99-100]. Compared to ML techniques, DL architectures offer several advantages, including high-accuracy, generalizability, DA and interpretability. Recent studies have demonstrated that DL architectures have achieved remarkable accuracy in diagnosing CVDs from CMR images [99-100][106]. These networks have the ability to learn patterns and complex relationships in CMRI data for early diagnosis of CVDs, which may not be easily identified by specialist doctors. Generalizability is another characteristic of DL networks compared to ML techniques. That is, DL techniques can be applied to CMR images with different registration protocols, yielding successful outcomes [106]. Additionally, various DA models based on DL techniques have been introduced and are currently utilized in the diagnosis of CVDs. For example, Shoeibi et al [20] utilized the CycleGAN technique in the pre-processing stage to detect myocarditis, achieving an accuracy rate of 99.71%. Interpretability is another capability of new DL techniques such as attention and transformer architectures [106]. This ability has made specialist doctors trust their results in the diagnosis of various diseases, including CVDs, in the near future.

Some DL models, such as CNNs, can provide visual explanations of their predictions, allowing clinicians to better understand the reasoning behind a diagnosis. In contrast, DL architectures face certain challenges, including overfitting and hardware resource requirements [53-59]. DL models can overfit the training data, meaning they may perform well on the training data but generalize poorly to new data [115-116]. Moreover, DL architectures demand significant hardware resources to facilitate rapid training, which may not be readily available to all researchers. In this section, the latest techniques of pertained and transformers are used in the diagnosis of myocarditis from CMR images. In this study, the pre-trained architectures include EfficientNet B3 [35], EfficientNet V2 [36], HrNet [37], Inception [38], ResNetrs50 [39], ResNest50d [40] and ResNet 50d [41]. These networks are among the latest pre-trained architectures that have achieved acceptable results in various applications. Therefore, we have utilized these models to diagnose myocarditis in our work.

Additionally, transformers architectures represent a new class of DL techniques that exhibit high efficiency in NLP applications and classification of medical images in comparison to conventional DL techniques [101-102]. In this study, the most important architectures of transformers have been tested to diagnose myocarditis, including Beit [42], Cait [43], Coat [44], Deit [45], Pit [46], Swin [47], TNT [48], Visformer [49] and ViT [50]. This section represents the second novelty of our paper, as it is the first study to examine and compare the techniques of transformers in diagnosing myocarditis from CMR images. This comparison is done with the aim of investigating the performance of the most important transformers techniques in diagnosing myocarditis from CMR images. In the following, detailed information of pre-trained architectures and transformers in the diagnosis of myocarditis from CMR images are provided. The following section provides a breakdown of the pre-trained model and transformer architectures, which are employed for detecting MCD through CMR images.

### 2.4.1. Deep Pre-Trained Models

The pre-trained models are a class of convolutional neural network (CNN) architectures based on supervised learning and has been trained on the ImageNet dataset [35]. Pre-trained models have emerged as a useful tool for researchers seeking to diagnose diseases as in medical field, researchers usually do not have access to large datasets related to the subject of interest [60-61]. In our study, we employed the last pre-trained models for MCD diagnosis, including EfficientNet B3 [35], EfficientNet V2 [36], HrNet [37], Inception [38], ResNetrs50 [39], ResNest50d [40], and ResNet 50d [41].

#### a) EfficientNet B3

In reference [35], the authors have comprehensively discussed the scaling of DL models. The research introduces a novel scaling approach that uniformly scales the depth, width, and resolution of models using compound coefficients. A few researchers have proposed a multi-objective approach to create DL models, which involves introducing the EfficientNets architecture [35]. This architecture is comprised of seven blocks, each of which contains various sub-blocks. Consequently, EfficientNets designs are identified with the suffixes B0 to B7, and further details can be found in [35].

#### b) EfficientNet V2

In reference [36], the authors proposed an integration of neuronal architecture search (NAS) and scaling to enhance DL models, specifically in terms of training speed and parameter efficiency. To achieve this, they constructed a search space comprising Fused-MBConv and utilized NAS to optimize model parameters such as accuracy, training speed, and size [36]. As a result, an 8-layer architecture, EfficientNetV2, was introduced, boasting a 4x faster training speed than the heart model [36]. Additionally, the size of these models was reduced by a factor of 6.8.

#### b) HrNet Model

The authors in [37] proposed an architecture called High-Resolution Net (HrNet), which demonstrates suitable performance in image classification. In study [37], the authors introduced a robust architecture named High-Resolution Net (HrNet), which has proven to be highly effective in image classification. Notably, recent research has also reported impressive results in the classification of grayscale images through the implementation of HrNet. The authors of [37] highlight that HrNet is a multi-resolution fusion-based model. There are two versions of HrNet available - namely, HRNetV1 and HRNetV2 [37]. An improved model of HRNetV2 called HRNetV2p was introduced with excellent performance [37].

#### c) Inception Models

Inception v4 is an upgraded version of the GoogLeNet architecture [38]. Compared to Inception v3, this architecture has a more homogenous architecture with higher efficiency. The Inception v4 consists of a stem block, three Inception blocks, two reduction blocks, and layers for average pooling/dropout/Softmax [38]. Given the success of residual networks, researchers [38] combined Inception and ResNets. They introduced two residual-based Inception models: Inception ResNet V1 and Inception ResNet V2 [38]. The architecture of these networks include: a stem block, three

Inception-ResNets, two reduction blocks, and the average pooling/dropout/Softmax layer [38]. Additional details about this model are presented in [38].

**d) ResNetrs50 (ResNet RS)**
Authors in [39] examined the impact of training methods and scaling strategies using ResNet architecture. In this work, an improved training procedure was presented through an experimental investigation of training methods without modification of the model architecture [39]. Analysis of scaling strategies presented two new methods: (i) overfitting can cause model depth scaling [39] and (ii) slower scaling of image resolution compared with other previous models [39]. Using this improved training and two scaling strategies, ResNet architecture was rescaled, which is known as ResNet-RS. This new architecture uses less memory during training and runs faster than its predecessor.

**e) ResNest50d (ResNeSt)**
In [40], Researchers proposed a simple architecture by combining the concepts of channel-wise attention strategy and multipath networks. This architecture captures cross-channel feature correlations in meta structure by maintaining independent representations [40]. In this architecture, a module executes a sequence of low-dimensional embedding transformations, combining their outputs into a multipath network [40]. Several variables can parameterize the architecture of this network. This computational block is called a Split-Attention Block [40]. Stacking the Split-Attention Block with ResNet-style creates a new type of ResNet called a Split-Attention network (ResNeSt) [40].

**f) ResNet 50d**
ResNet uses skip connection or Residual Connections in its architecture [41]. The technique of incorporating the output of the previous layer into the input of the subsequent layer, known as skip connections, has been shown to effectively mitigate the issue of vanishing gradients in deep neural networks. Through this approach, certain layers within the network can be skipped, resulting in improved performance and convergence rates. The ResNet family comprises various architectures, including ResNet-18, ResNet-34, ResNet-50, ResNet-101, and ResNet-152 (numbers indicate the number of layers in each architecture) [41]. More information about the ResNet architecture can be found in reference [41].

### 2.4.2. Deep Transformer Models
Recent works suggest that it is also possible to build successful classification models without convolutions [62], and a new family of image recognition models called transformer architecture are used. Transformers [63] were initially designed to tackle natural language processing (NLP) tasks by introducing a self-attention module that captures the non-local relationships between all input sequence elements. However, it is also used in computer vision by seeing an image as a sequence of patches and following the architectural design [63]. Figure (6) displays a general block diagram for transformer models. There are several transformer models used for computer vision. They are usually pre-trained on the ImageNet dataset as convolutional models.

**a) Beit**
Beit [42] is a transformer architecture that follows the design of [63], and trained in a self-supervised manner following the ideas of the language model BERT [42]. A masked image modelling (MIM) pre-training task is proposed which uses two views for each image which are image patches and visual tokens. Each image is split into a grid of patches that are the input representation of the transformer architecture. Moreover, the image is tokenized into discrete visual tokens. During pre-training, some proportion of image patches are masked, and the corrupted input is fed to the transformer. Using this approach, the model learns to recover the visual tokens of the original image. After this pre-training task, a fine-tuning process for image classification is conducted using the representations learnt during the pre-training process.

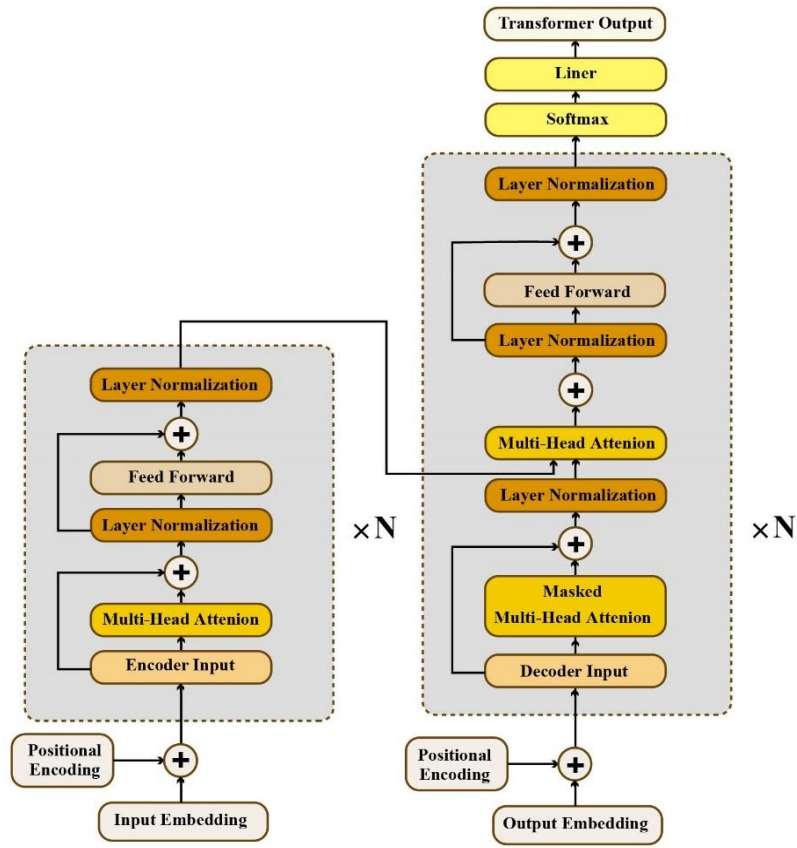

Fig. 6. General block diagram for deep transformer models.

**b) Cait**

Cait [43] architecture was designed to increase the optimization's stability when training transformers for image classification derived from the original transformer architecture by [63], especially when the depth of the architecture is increased. In the Cait architecture, two improvements were introduced: a new layer called Layer Scale, which allows the authors to train deeper high-capacity image transformers that benefit from depth; and a class-attention layer is devoted to extract the content of the processed patches into a single vector so that it can be fed to a linear classifier.

**c) Coat**

Coat [44] is a transformer-based image classifier with co-scale and conv-attentional mechanisms. The co-scale mechanism maintains the integrity of transformers' encoder branches at individual scales while allowing representations learned at different scales to communicate with each other effectively. The conv-attentional mechanism provides a relative position embedding formulation with an efficient convolution-like implementation.

**d) Deit**

In general, transformers do not generalize well when trained on insufficient data, this drawback was addressed by [45]. The Deit architecture is based on the Vit architecture [50] but uses a training that considers strategies for hyper parameter optimization, DA, and regularization, allowing the authors to learn vision transformers in a data-efficient manner. Moreover, a teacher-student strategy that is specific for training transformers was also proposed in this work.

**e) Pit**

The authors of Pit [46] proposed a pooling-based vision transformer using the original ViT model [50]. The Pit architecture is a transformer architecture combined with a pooling layer that enables the spatial size reduction in the ViT structure as in the ResNet convolutional network [46]. This architecture tries to mimic the approach followed in CNNs, which starts with a feature of large spatial sizes and a small channel size which gradually increases the channel size while decreasing the spatial size.

**f) Swin**
The Swin [47] transformer was designed to address the challenges in adapting transformers from language to vision due to the differences in these two domains. The differences are mainly the large variations in the scale of visual entities and the high resolution of pixels in images compared to words in text. To address these differences, the Swin architecture is a hierarchical transformer whose representation is computed with shifted windows is employed.

**g) Turbulence Neural Transformer (TNT)**
Vision transformers first divide the input images into several local patches and then calculate both representations and their relationship. However, due to the high complexity with abundant details and color information of natural images, the granularity of the patch dividing is not fine enough for excavating features of objects in different scales and locations. In the TNT architecture [48], the local patches are seen as "visual sentences," and they are further divided into smaller patches as "visual words". Features of both visual words and sentences are aggregated to enhance the representation ability of this architecture.

**h) Visformer**
The authors of Visformer [49] deal with the over-fitting problem in vision transformers when the training data is limited. To this aim, the authors studied the transition from the Deit transformer model [45] to the ResNet convolutional network [49]. This architecture uses the advantages of transformers and discards the disadvantages of obtaining a model that outperforms both Deit and ResNet.

**i) ViT**
Vit [50] was the first fully-transformer network for image classification. In this work, images were split into fixed-size patches; subsequently, those patches were linearly embedded, combined with positional embedding, and finally fed to a standard transformer encoder that follows the design of [63]. An extra learnable "classification token" is added to the sequence of linearly embedded patches to perform classification. The Vit models were trained by combining three datasets: ILSVRC-2012 ImageNet, ImageNet-21k, and JFT.

**3. Statistical Metrics**
This section presents the evaluation metrics for demonstrating the efficacy of our proposed method. A 10-fold cross-validation was used to evaluate the classification results. The advantage of the 10-fold method is its application to train and experiment data. Evaluation metrics, including specificity (Spec), sensitivity (Sens), accuracy (Acc), and F1 score, were calculated for each of the DL models. Table 1 shows the formula behind each evaluation metric. For each evaluation metric, true positives (TP), false negatives (FN), true negatives (TN), and false positives (FP) are obtained from the confusion matrix [64].

Table 1. Description of the performance metrics used in this work.

| Parameter Name | Formula |
| --- | --- |
| Accuracy (Acc) | $Acc = \dfrac{TP + TN}{FP + FN + TP + TN}$ |
| Sensitivity (Sens) | $Sens = \dfrac{TP}{FP + TP}$ |
| Specificity (Spec) | $Spec = \dfrac{TN}{FP + TN}$ |
| Precision (Prec) | $Prec = \dfrac{TP}{TP + FP}$ |
| F1-score (FS) | $FS = \dfrac{2\,TP}{2\,TP + FP + FN}$ |

## 4. Experiment Results

This section presents the results obtained for MCD diagnosis with pre-trained and transformed models. All simulations were run on a hardware system with 16 GB RAM, CPU Core i7, and NVidia GeForce 2080. Also, the proposed CADS was implemented using TensorFlow 2 [73], Keras [74], FastAI [52], and Scikit-learn [75], and PyTorch [76] tools. In section 2.2, we report the results of the proposed preprocessing methods for diagnosing MCD in CMR images. In this section, we first perform pre-processing steps, including noise removal and dimension reduction for CMR images, and then provide detailed explanations about them. Our work aims to enhance the efficiency of the proposed CADS for MCD diagnosis using DL techniques, as previously discussed. In the continuation of pre-processing, CutMix and MixUp techniques [32-33] are utilized for DA. In figure (5) the results of the proposed DA technique are displayed. As can be seen, the proposed DA method has proven to be highly effective in generating artificial CMR images. Furthermore, in this section, the results of pretrained techniques and proposed transformers for MCD diagnosis are presented.

In the beginning of this section, the results of pre-trained models are discussed, including EfficientNet B3 [35], EfficientNet V2 [36], HrNet [37], Inception [38], ResNetrs50 [39], ResNest50d [40], and ResNet 50d [41]. These networks belong to the new category of pre-trained models, which have been utilized for various applications and have demonstrated satisfactory result. Consequently, in this experiments section, these pre-trained models are employed to diagnose myocarditis from CMR images. Table (2) presents the hyper-parameters of the pre-trained models, including batch size, learning rate, epochs, and size of input images.

Table 2. The hyper-parameters for pretrained models in MCD detection

| Model | Batch Size | Learning Rate | Epochs | Image Size |
|---|---|---|---|---|
| resnetrs50 | 128 | 3e-3 | 100 | 224 |
| inception_v4 | 128 | 3e-3 | 100 | 224 |
| inception_resnet_v2 | 128 | 3e-3 | 100 | 224 |
| resnest50d | 128 | 3e-3 | 100 | 224 |
| resnet50d | 128 | 3e-3 | 100 | 224 |
| efficientnetv2_rw_s | 64 | 3e-3 | 100 | 224 |
| efficientnet_b3 | 64 | 3e-3 | 100 | 224 |
| hrnet_w32 | 64 | 3e-3 | 100 | 224 |

Table (3) shows the results of the pre-trained models for MCD diagnosis using CMRI images. At the onset of this section, the results of various pre-trained models, without the proposed DA method, for the diagnosis of myocarditis are displayed. As can be seen, the mean of the evaluation parameters is reported alongside with their variance for each of the pre-trained models. Table (3) shows that the Inception ResNet v2 model was more successful in MCD diagnosis than other pre-trained models. The ResNet 50d model also achieved good results. The accuracy results for pre-trained models in diagnosing MCD are illustrated in Figure (7).

Table 3. Results (%) of pertained models for diagnosis of MCD from CMRI images

| Models | ACC | Prec | NPY | Rec | Spec | F1 | AUROC | Cohen |
|---|---|---|---|---|---|---|---|---|
| EfficientNet B3 | 99,52 (3,48) | 99,47 (17,87) | 99,55 (6,43) | 99,14 (30,96) | 99,72 (17,85) | 99,3 (24,07) | 99,96 (8,15) | 98,94 (13,9) |
| EfficientNet V2 | 99,54 (2,58) | 99,19 (15,64) | 99,72 (4,39) | 99,47 (20,46) | 99,57 (6,89) | 99,33 (21,2) | 99,98 (7,97) | 98,98 (13,95) |
| HrNet | 99,04 (0,28) | 98,11 (0,31) | 99,54 (0,27) | 99,12 (0,51) | 99 (0,16) | 98,61 (0,41) | 99,95 (0,05) | 97,88 (0,63) |
| Inception ResNet V2 | 99,66 (0,18) | 99,31 (0,49) | 99,85 (0,1) | 99,71 (0,18) | 99,64 (0,26) | 99,51 (0,26) | 99,99 (0,01) | 99,26 (0,4) |
| Inception V4 | 99,30 (0,32) | 98,46 (1,08) | 99,75 (0,34) | 99,52 (0,66) | 99,18 (0,59) | 98,99 (0,46) | 99,98 (0,01) | 98,45 (0,71) |
| ResNest 50d | 99,23 (0,4) | 99,69 (0,44) | 98,99 (0,37) | 98,06 (0,72) | 99,84 (0,23) | 98,87 (0,58) | 99,93 (0,09) | 98,28 (0,88) |
| ResNet 50d | 99,62 (0,14) | 99,51 (0,27) | 99,68 (0,2) | 99,39 (0,38) | 99,74 (0,14) | 99,45 (0,2) | 99,99 (0,01) | 99,16 (0,3) |
| ResNetrs 50 | 99,42 (0,33) | 99,04 (0,69) | 99,63 (0,54) | 99,29 (1,03) | 99,49 (0,37) | 99,16 (0,48) | 99,96 (0,08) | 98,72 (0,72) |

In the following, the results of the pre-trained model along with a proposed DA method based on the CutMix [32] and MixUp [33] techniques are presented. To the best of our knowledge, this study is the first study to use this combination for the diagnosis of MCD. The proposed DA method generated an artificial CMRI image, which was used as input for the pre-trained model. Table (4) shows the performance of the pre-trained architectures achieved using the new DA method. The inception_v4 model performed better than other pre-trained architectures, while the inception of ResNet V2 and efficientnetv2 architectures also achieved satisfactory results. Figure (7) shows the accuracy results for pre-trained models combined with the DA method for the diagnosis of MCD.

Table 4. Results (%) obtained using pertained models with the new DA method for diagnosis of MCD with CMR images

| Models | ACC | Prec | NPY | Rec | Spec | F1 | AUROC | Cohen |
|---|---|---|---|---|---|---|---|---|
| EfficientNet B3 | 99,55 (0,27) | 99,23 (0,44) | 99,72 (0,25) | 99,47 (0,47) | 99,59 (0,23) | 99,35 (0,39) | 99,97 (0,03) | 99,01 (0,6) |
| EfficientNet v2 | 99,66 (0,22) | 99,35 (0,48) | 99,83 (0,16) | 99,67 (0,31) | 99,66 (0,25) | 99,51 (0,33) | 99,95 (0,06) | 99,25 (0,5) |
| HrNet | 99,10 (0,52) | 97,98 (1,76) | 99,72 (0,25) | 99,47 (0,47) | 98,91 (0,96) | 98,71 (0,74) | 99,95 (0,05) | 98,02 (1,14) |
| Inception ResNet v2 | 99,68 (0,24) | 99,47 (0,49) | 99,79 (0,21) | 99,59 (0,41) | 99,72 (0,26) | 99,53 (0,34) | 99,99 (0,02) | 99,29 (0,52) |
| Inception v4 | 99,70 (0,18) | 99,44 (0,3) | 99,84 (0,14) | 99,69 (0,26) | 99,71 (0,16) | 99,57 (0,25) | 99,94 (0,1) | 99,34 (0,39) |
| ResNest 50d | 99,33 (0,3) | 99,02 (0,52) | 99,49 (0,46) | 99,02 (0,89) | 99,49 (0,28) | 99,02 (0,44) | 99,94 (0,06) | 98,51 (0,67) |
| ResNet 50d | 99,58 (0,19) | 99,47 (0,23) | 99,64 (0,36) | 99,31 (0,7) | 99,72 (0,12) | 99,39 (0,28) | 99,99 (0,02) | 99,07 (0,43) |
| ResNetrs 50 | 99,43 (0,18) | 98,75 (0,62) | 99,79 (0,13) | 99,59 (0,25) | 99,34 (0,33) | 99,17 (0,26) | 99,98 (0,02) | 98,73 (0,4) |

The rest of this section presents the results of the transformer models for MCD diagnosis. Recently, transformer models have been introduced and have been successfully used in various medical imaging [15-17]. In this work, the most recent transformer models have been investigated and compared for MCD diagnosis using CMRI images. In this paper, the transformer models that have been utilized are Beit [42], Cait [43], Coat [44], Deit [45], Pit [46], Swin [47], TNT [48], Visformer [49], and ViT [50]. In table (5), hyper-parameters of transformers models including batch size, learning rate, epochs and input images size are displayed.

Table 5. The hyper-parameters for transformer models in MCD detection

| Model | Batch Size | Learning Rate | Epochs | Image Size |
|---|---|---|---|---|
| Beit | 32 | 3e-3 | 250 | 224 |
| Cait | 16 | 3e-3 | 250 | 224 |
| Coat | 8 | 2e-4 | 250 | 224 |
| Deit | 32 | 3e-3 | 250 | 224 |
| Pit | 16 | 3e-3 | 250 | 224 |
| Swin | 16 | 3e-5 | 250 | 224 |
| TNT | 16 | 1e-4 | 250 | 224 |
| Twins | 32 | 2e-4 | 250 | 224 |
| ViT | 32 | 1e-4 | 250 | 224 |
| Visformer | 32 | 2e-4 | 250 | 224 |

The transformer models' outcomes for MCD diagnosis are presented in Table (6). The results indicate that the Coat and TNT models have demonstrated effective outcomes for diagnosing MCD from CMRI images. Figure (8) displays the accuracy outcomes of transformer models for MCD diagnosis. Based on Figure (8), it is apparent that the efficiency of transformer models, such as Coat [44], Swin [47], TNT [48], Visformer [49], and ViT [50], is enhanced through the utilization of DA, in contrast to when it is not employed. However, the use of DA has resulted in reduced performance of Beit [42], Cait [43], Deit [45], and Pit [46] architectures, which are not well trained using small input data. Consequently, applying more input data using the proposed DA technique to the input of these networks has resulted in diminished evaluation parameters. These findings highlight that not all transformer models yield satisfactory outcomes in diagnosing myocarditis from CMR images.

Table 6. Results (%) obtained using transformer models for diagnosis of MCD with CMRI images.

| Models | Acc | Prec | NPY | Rec | Spec | F1 | AUROC | Cohen |
|---|---|---|---|---|---|---|---|---|
| Beit | 75,98 (4.26) | 65,32 (7.23) | 82,23 (3.59) | 65,86 (7.49) | 81,26 (6.63) | 65,09 (3,75) | 82,01 (3,74) | 46,88 (3.38) |
| Cait | 79,30 (4.01) | 68,24 (11,34) | 86,02 (4.78) | 74,08 (11,41) | 82,03 (9,96) | 70,89 (6,04) | 87,43 (5.02) | 54,89 (7.11) |
| Coat | 99,68 (0,27) | 99,55 (1.52) | 99,74 (0,49) | 99,51 (1.24) | 99,77 (0,79) | 99,53 (0,39) | 99,68 (0,097) | 99,28 (0,83) |
| Deit | 84,68 (0,97) | 83,12 (4.34) | 85,40 (3.07) | 69,66 (6.09) | 92,53 (4,17) | 75,72 (2,71) | 91,15 (2.25) | 64,66 (3.23) |
| Pit | 97,41 (3.03) | 95,02 (6.07) | 98,76 (1,09) | 97,67 (3.38) | 97,27 (3,16) | 96,30 (3.08) | 98,96 (1.49) | 94,31 (2,98) |
| Swin | 96,90 (3.17) | 96,27 (4.39) | 97,22 (1,14) | 94,61 (2.69) | 98,10 (2.41) | 95,43 (2,16) | 98,34 (1.38) | 93,09 (4.37) |
| TNT | 99,68 (0,41) | 99,47 (0,58) | 99,79 (0,16) | 99,59 (0,33) | 99,72 (0,16) | 99,53 (0,61) | 99,94 (0,16) | 99,28 (0,59) |
| Visformer | 98,23 (0,92) | 96,63 (0,78) | 99,15 (1,23) | 98,37 (2.48) | 98,16 (0,57) | 97,46 (0,12) | 99,43 (0,86) | 96,11 (2.03) |
| ViT | 75,98 (3.68) | 65,32 (6,57) | 82,23 (3,31) | 65,86 (7.73) | 81,26 (6,98) | 65,09 (4,09) | 82,01 (4.58) | 46,88 (3.14) |

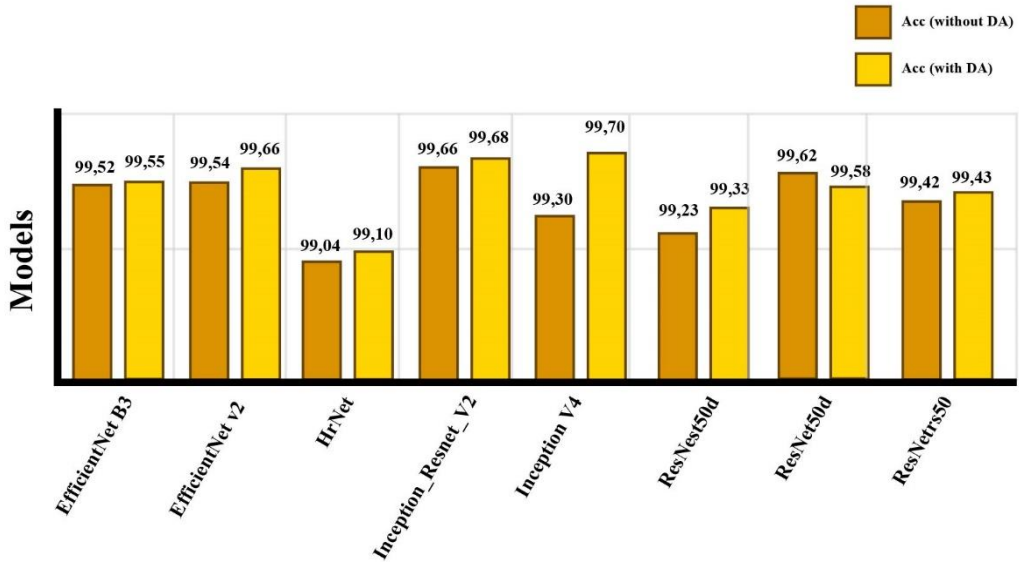

Fig. 7. Accuracy (%) obtained using various pre-trained models.

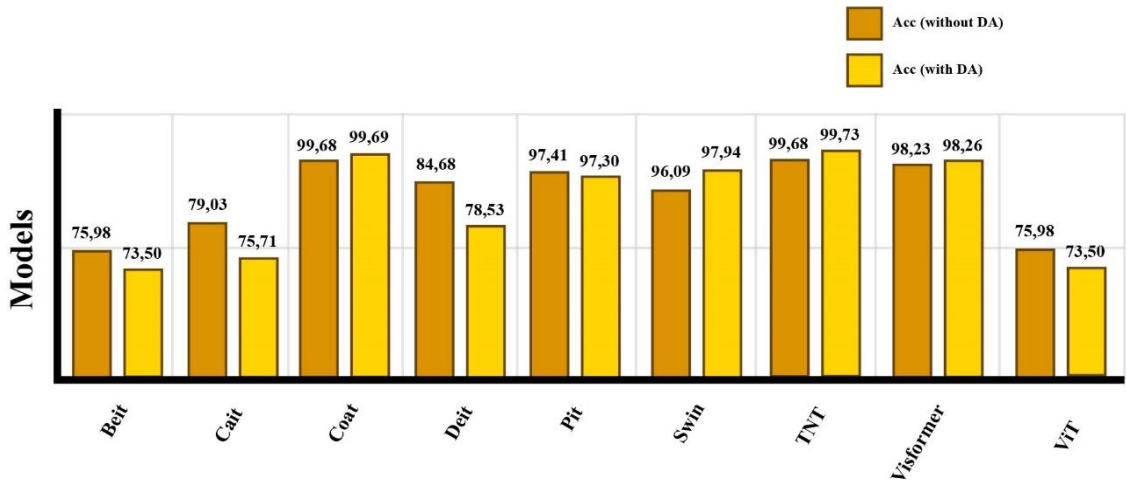

Fig. 8. Accuracy (%) obtained using various transformer models.

Finally, the results (%) of the transformers models trained with the new DA method are provided in Table (7). The proposed DA method generates artificial CMRI images, which boosts the performance of CADS in detecting MCD. The results of the transformer architecture combined with the proposed DA method are presented in Table (7). As evident from the table, the utilization of DA technique and transformer architecture has significantly enhanced the efficiency of the CADS proposed for MCD diagnosis. Table (7) displays that the TNT architecture has outperformed other transformers and pre-trained techniques. Figure (8) illustrates the comparison of accuracy outcomes for transformer models with and without the implementation of the proposed DA method for MCD diagnosis. Furthermore, ROC curves for transformers with DA are illustrated in Figures (9) and (10).

Table 7. Results (%) of transformer models with new DA for diagnosis of MCD from CMRI images

| Models | Acc | Prec | NPY | Rec | Spec | F1 | AUROC | Cohen |
|---|---|---|---|---|---|---|---|---|
| Beit | 73,5 (2,25) | 64,3 (6,73) | 78,33 (2,19) | 55,74 (8,66) | 82,78 (6,3) | 58,89 (3,04) | 77,17 (3,37) | 39,63 (2,46) |
| Cait | 75,71 (3,66) | 69,41 (10,18) | 79,81 (3,57) | 58,02 (12,82) | 84,95 (9,55) | 61,69 (5,98) | 81,48 (4,03) | 44,38 (6,93) |
| Coat | 99,69 (0,18) | 99,39 (0,61) | 99,65 (0,27) | 99,32 (0,51) | 99,68 (0,32) | 99,35 (0,26) | 99,89 (0,09) | 99,02 (0,39) |
| Deit | 78,53 (0,94) | 70,53 (3,26) | 82,45 (1,69) | 64,97 (5,32) | 85,62 (3,22) | 67,43 (2,04) | 85,13 (1,31) | 51,49 (2,13) |
| Pit | 97,3 (2,13) | 94,12 (4,84) | 99,18 (0,9) | 98,47 (1,65) | 96,69 (2,78) | 96,21 (2,95) | 99,1 (0,53) | 94,12 (4,61) |
| Swin | 97,94 (1,77) | 96,72 (3,25) | 98,6 (0,97) | 97,35 (1,81) | 98,25 (1,77) | 97,03 (2,52) | 99,46 (0,64) | 95,45 (3,88) |
| TNT | 99,73 (0,17) | 99,19 (0,45) | 99,89 (0,08) | 99,8 (0,14) | 99,57 (0,24) | 99,49 (0,25) | 99,94 (0,07) | 99,22 (0,38) |
| Visformer | 98,26 (0,63) | 96,78 (0,69) | 99,06 (0,7) | 98,2 (1,35) | 98,29 (0,36) | 97,48 (0,92) | 99,4 (0,53) | 96,16 (1,4) |
| ViT | 73,5 (2,25) | 64,3 (6,73) | 78,33 (2,19) | 55,74 (8,66) | 82,78 (6,3) | 58,89 (3,04) | 77,17 (3,37) | 39,63 (2,46) |

## 5. Post Processing using Explainable AI

One of the main drawbacks of machine learning (ML) and DL classification models is their black-box nature, which hinders the usage and trust of these models. An approach to tackle this issue is the application of model interpretability techniques [65]. Interpretable machine learning methods extract relevant knowledge from a machine learning model concerning relationships either contained in data or learned by the model [66]. A particular case of interpretability methods is explainability techniques, which provide explanations of individual predictions of a given model [67].

Following the taxonomy [68], we distinguish two kinds of interpretability methods: intrinsic and post hoc. Intrinsic methods refer to machine learning models considered interpretable due to their simple structure, such as short decision trees or sparse linear models. The post hoc interpretability refers to applying interpretation methods after model training. Due to the nature of deep learning models, interpretability methods for deep learning techniques are mainly post hoc methods. In the context of interpretability methods for deep image classification models, we can find several techniques, such as LIME [69], saliency maps [70], or integrated gradient attribution [71].

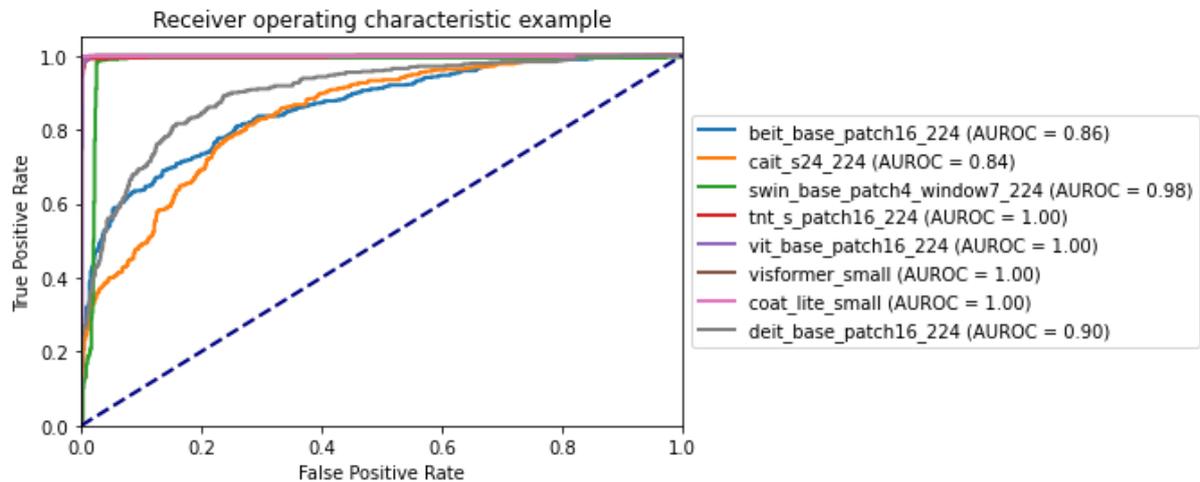

Fig. 9. ROC curves for transformer models in the diagnosis of MCD using CMRI images.

Among the available XAI techniques for image classification models, we employed the occlusion-based attribution algorithm [72], supported by the Captum library [73]. Using this algorithm, we estimated areas of the image that are critical for the classifiers' decision by occluding them and quantifying how the decision changed. We run a sliding window of size $15 \times 15$ with a stride of 8 along both image dimensions. At each location, we occluded the image with a baseline value of 0, corresponding to a black patch.

The results of the Grad-Cam method applied to both the best pre-trained and transformer models for diagnosing MCD are illustrated in Figure (11). The highlighted regions in the CMRI images indicating potential abnormalities in MCD could be successfully identified. This post-processing step helps cardiologists to diagnose MCD in the early stages which is one of the novelties of this paper.

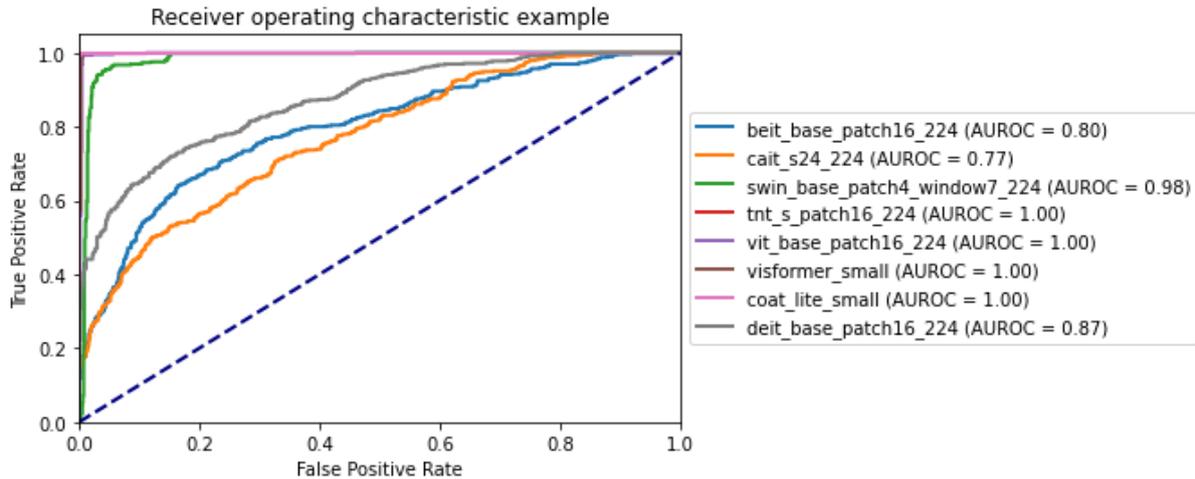

Fig. 10. Roc curves for transformer models trained with new DA techniques in MCD detection from CMR images.

## 6. Limitations of Study

The experiments detailed in the previous section were performed exclusively on the Z-Alizadeh dataset, which comprises 5538 CMRI images of MCD patients and 5041 HC images. However, the dataset may not be ideal for training advanced DL models primarily because of the limited number of CMR images of MCD and HC available. In addition, this dataset does not cover the full range of severity levels for MCD diagnosis, which presents a major challenge for predicting disease severity using DL models. One possible solution would be to incorporate multispectral data such as X-rays images alongside CMRI to generate more informative datasets for MCD diagnosis. It is imperative to note that the Z-Alizadeh dataset is designed specifically for diagnostic purposes and not severity predictions. Furthermore, it comprises only two classes (MCD and HC), but in future applications, a multi-class DL model may be utilized to diagnose different types of CVDs.

As mentioned in the previous sections, the use of transformers architectures is one of the most important novelties of this work [101-102]. These networks offer the advantage of high accuracy and efficiency with low input data. However, they necessitate high-power hardware resources for proper training, which are typically unavailable to the general public. This challenge has led to prolonged training times for transformer architectures on CMRI data. To overcome this challenge in the future, compression techniques of DL models such as quantization methods [117] can be employed to enable training of transformer networks on medium-power hardware in less time. In another part of this paper, Grad-Cam method is utilized as an XAI technique [107-108], which can assist specialist physicians in diagnosing myocarditis from CMR images. Nonetheless, using these methods in conjunction with DL techniques can be complex. Hence, it is anticipated that XAI approaches will be increasingly employed in the precise diagnosis of myocarditis in the future.

## 7. Discussion,

Myocarditis is an inflammatory condition that affects the heart muscle and is caused by various factors including viral infections, bacterial infections and autoimmune disorders [19-22]. CMRI is a non-invasive method for diagnosing myocarditis, but this medical imaging method can pose challenges for doctors [29-31]. Symptoms and findings of myocarditis frequently overlap with those of other CVDs, such as cardiomyopathy [118], ischemic heart disease [119], and arrhythmogenic right ventricular cardiomyopathy [120]. As a result, distinguishing myocarditis from these other conditions based solely on MRI findings can be challenging. Additionally, CMR images may be associated with various artifacts and noises that make the diagnosis of myocarditis challenging. Consequently, the raised challenges have led to the presentation of AI techniques including ML and DL methods in the diagnosis of CVDs [97-100].

In this study, a DL-based CADS is proposed to diagnose myocarditis from CMR images. Figure (1) shows the block diagram of the proposed method, which includes the dataset, pre-processing, DL model and post-processing sections. As stated in the introduction section, so far limited research has been conducted in the field of myocarditis diagnosis from CMR images using AI techniques [97-100]. Notably, the present study offers a novel approach that has never been presented before, featuring three key novelties in the pre-processing, classification, and post-processing stages.

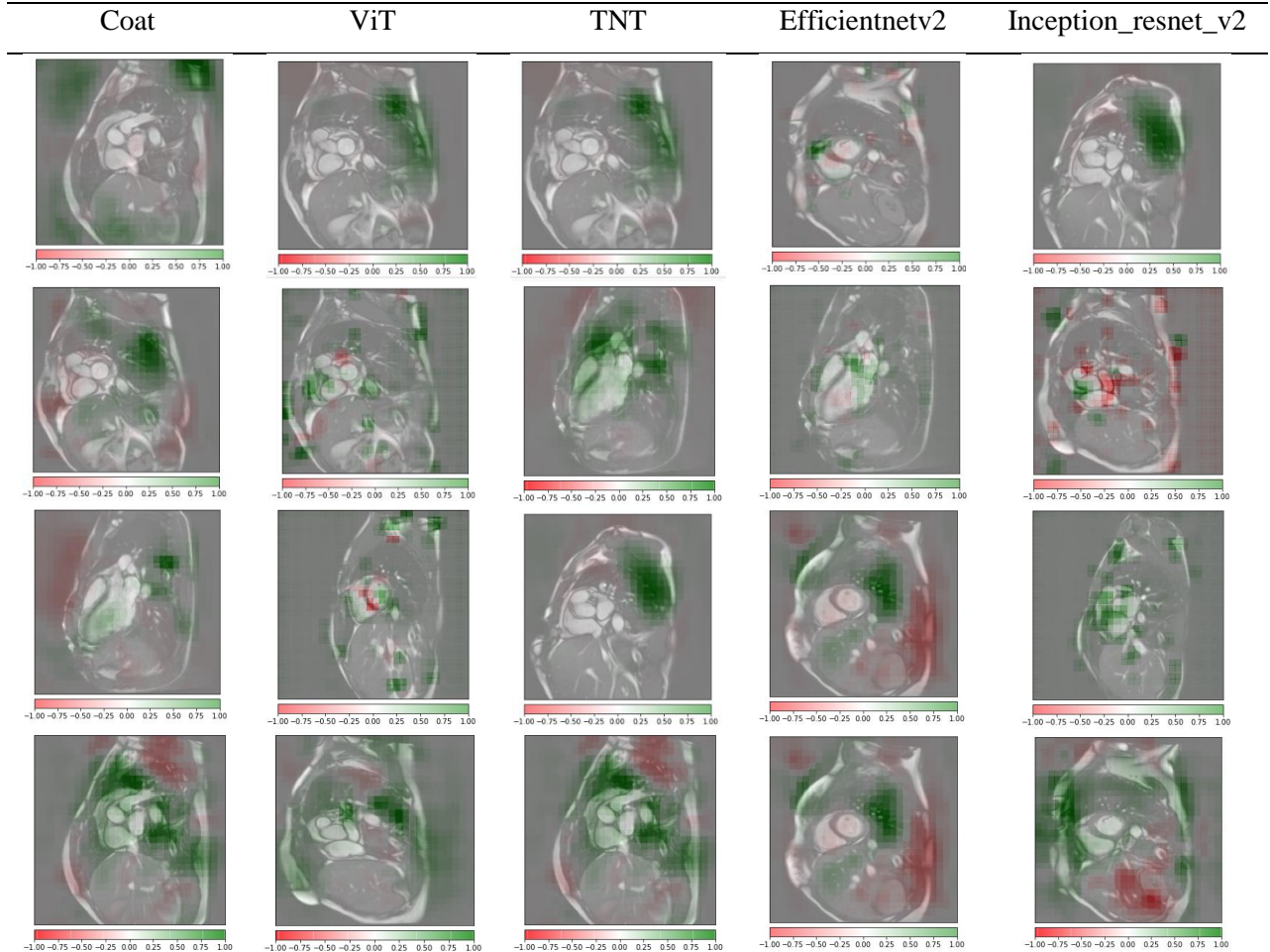

Fig. 11. Plot Grad-Cam for the best transformer and pre-trained models based on the new DA method

In order to perform the experiments, the Z-Alizadeh dataset [19] is utilized, which has been registered by some researchers of this paper. Z-Alizadeh dataset consists of ludes 12000 CMRI data from HC and MCD patients, which were registered in Shahid Rajaee Hospital in Tehran. Preprocessing is performed on the dataset, including denoising, resizing, and the development of a new DA method [32-33] for generating artificial CMR images. As some images in the dataset exhibit low contrast and various artifacts, a Gaussian filter was applied to mitigate these issues. Given that doctors record CMR images from diverse regions of the heart, so the dataset images have different dimensions that must be resized to one size. Therefore, in the second stage of pre-processing, all images were resized to 224x224 size. Furthermore, artificial CMRI data are generated using a new DA model based on CutMix [32] and MixUp [33] techniques. Prior research [97-100][106] has not employed these techniques in the pre-processing step in CVDs detection research. The proposed DA method is used for the first time in MCD diagnosis and is the first novelty in this paper. The results of the DA method, presented in Figure (5), demonstrate its effectiveness in generating artificial CMR images.

In the third step, various pretrained models and transformers were used to extract features and classify CMR images to diagnose myocarditis. Deep transfer learning models have shown great promise in the classification of medical images [121-122]. By leveraging pre-trained CNN models and fine-tuning them on medical image datasets, these models can learn to extract relevant features from medical images and improve classification performance [121]. Transfer learning has the potential to enable the development of accurate and reliable medical image classification models, which can aid in the diagnosis and treatment of various medical conditions [122]. However, it is important to note that transfer learning is not a one-size-fits-all solution and the choice of pre-trained CNN model, the size and quality of the medical image dataset, and the task-specific layer architecture can all impact the performance of the model [121-122]. Therefore, careful consideration and evaluation of different transfer learning techniques are necessary to develop accurate and reliable medical image classification models [121]. In this work, Pre-trained models for MCD diagnosis from CMR images included EfficientNet B3 [35], EfficientNet V2 [36], HrNet [37], Inception [38], ResNetrs50 [39], ResNest50d [40], and ResNet 50d [41].

Deep transformer models have also shown great potential in the classification of medical images [101]. The Transformer architecture was originally proposed for natural language processing tasks, but it has been adapted for computer vision tasks, including image classification [102]. In medical image classification, the deep transformer model takes the image as input and generates a feature representation using multiple layers of self-attention and feedforward neural networks [101-102]. The feature representation is then passed through a final classification layer to produce the output class probabilities [102]. Medical images such as CMRI often have different sizes and aspect ratios, and traditional CNNs require the images to be resized to a fixed size before feeding them to the model. Deep transformer models can handle variable-sized inputs without the need for resizing, which can preserve important details in the images [102]. However, one potential challenge with deep transformer models is their computational complexity. The transformer models also included Beit [42], Cait [43], Coat [44], Deit [45], Pit [46], Swin [47], TNT [48], Visformer [49], and ViT [50]. The transformer model for diagnosing MCD is another novelty of this work.

In comparison to prior studies [97-100], it is evident that so far in all the researches of CVDs diagnosis, a comprehensive comparison between the new transformer-based methods has not been conducted. This section will aid researchers in utilizing transformers models in future research in diagnosing other CVDs based on their performance. In the experiments section, the results of the proposed method are reported in tables (6) and (7). According to the table (7) and figure (8), it can be seen that TNT transformer model is able to achieve 99.71% accuracy. It can also be seen in the figure (8) and table (7) that the utilization of the proposed DA technique has resulted in a reduction accuracy for some transformers models. The reason for this is that these methods are not fully trained on limited data. Accordingly, when exposed to large datasets, their actual performance is revealed.

In the following, the proposed method in this paper incorporates the Grad-Cam technique as a post-processing approach [107-108]. The Grad-Cam method is one of the most important XAI techniques that has recently gained attention in the diagnosis of various diseases, including brain tumors and CVD diseases [107-108]. This technique has the potential to assist specialist doctors in facilitating the rapid diagnosis of diseases. In this work, Grad-Cam method is used as a post-processing method for the best pretrained and transformer models. In figure (11), the results of the Grad-Cam method for diagnosing myocarditis from CMR images are displayed. Based on the results presented in Figure (11), the Grad-Cam technique for the Coat model shows successful performance. Also, this method has shown successful results for ViT, TNT, Efficientnetv2 and Inception_resnet_v2 techniques to diagnose myocarditis. This section is the third novelty in this paper. In the references [97-100], prior research on the diagnosis of CVDs from CMR images using ML and DL techniques is presented. It can be seen that XAI techniques have not been utilized in the diagnosis of myocarditis and most CVD diseases.

## 8. Conclusion, and Future Works

The myocardium is the most critical part of the heart which is responsible for contracting, pumping, and distributing blood throughout the body. Damage to myocardial tissue leads to MCD, which causes many complications for patients due to devastating effects on the heart muscle and electrical system [19]. This disease significantly impairs the heart's ability to pump blood, leading to arrhythmias. MCD also puts individuals at risk of developing blood clots within the heart, which can result in stroke or heart attack [20-21]. Various factors such as viral and bacterial infections, heart surgery, rheumatic fever, as well as certain drugs and toxins [19-22], lead to the development of MCD in individual.

Until now, multiple screening methods have been introduced by specialist doctors to diagnose CVDs [22-28]. CMR images provide important information regarding heart tissue, which physicians use for CVD diagnosis in the early steps [29]. Despite the benefits of CMR imaging in diagnosis of CVDs, it comes with some challenges that medical experts need to address. For example, MCD diagnosis mandates high-slice CMR images [29-31]. Analyzing high-slice CMR images can be time-consuming for specialist doctors [31]. Medical professionals may sometimes incorrectly diagnose MCD, which may be influenced by factors like eyestrain and various artifacts in CMR images. To address these challenges, several research has been conducted to diagnose CVDs such as early detection of MCD using DL models [16-19].

This paper proposes a new DL-based CADS for MCD diagnosis from CMR images. The proposed method involves several steps, including dataset, preprocessing, feature extraction and classification using DL models, and post-processing. First, the Z-Alizadeh Sani myocarditis dataset was used in this work. In the preprocessing step, denoising, resizing, and DA also was employed to generate an artificial CMRI images. In the third step, the latest deep pre-trained and transformers models are used in feature extraction and classification steps for diagnosis of MCD. Experiment results indicated that combining the TNT architecture with the proposed DA method help to achieve the highest performance. The XAI based grad cam method also was used in the post-processing step to indicate MCD suspicious regions in CMR images. This section implemented the grad cam method on pre-trained and transformer models. Table (8) compares the results of the proposed method with various research works done on automated CVD detection.

Table 8. Comparison of the proposed method with related works.

| Ref | Application | Dataset | Preprocessing | DNN | Performance (%) |
|---|---|---|---|---|---|
| [80] | Myocardial | MICCAI 2009 | ROIs Extraction | CNN | Acc=86.39 Sen=90 |
| [81] | End-Diastole and End-Systole Frames | Free-Breathing CMR Data STACOM2011 | DA | 2D-CNN | Acc= 76.5 |
| [82] | Heart and right ventricle | Clinical York University | DA | NF-RCNN | AUC: 0.98 recall: 0.96 |
| [83] | Detection | Shenzhen Maternal and Child Health Hospital | ROI detection | 2D-CNN | Acc=94.84 Sen=92.73 Spec=94.27 |
| [84] | Classification of MYO Delayed Enhancement Patterns | Clinical | -- | GoogLeNet, AlexNet, ResNet-152 | Acc=79.5 |
| [85] | Classification and Prediction | Clinical | -- | deeplabV3 InceptionResnetV2 | Different Results |
| [86] | Cardiac view | Clinical | -- | Autoencoder | Acc: 96.7 |
| [87] | LV | Clinical | ROI extraction | AE | Acc=:97.5 Sens:84.2 Spec:98.6 |
| [88] | MYO | Clinical | -- | CNNEC | Acc: 95.3 |
| [89] | Cardiac Contraction | UK Biobank | -- | cGAN | DM: 0.89 |
| [90] | Myocardial | MICCAI 2020 EMIDEC | Different Methods | U-Net | -- |
| [91] | Myocardial | Clinical | Standard Preprocessing | 2D-CNN | Acc= 96.70 |
| [92] | Myocardial | MICCAI 2020 EMIDEC | Normalization, Resampling, Segmentation, Applying Three-order Spline Interpolation | 2D U-Net | Acc= 92.00 |
| [93] | Myocardial | Clinical | Resampling, Cropping, CLAHE, LV Localization, Filtering Methods | ResNet-56 | -- |
| [94] | Myocardial | Clinical | Standard Preprocessing | GoogLeNet, AlexNet, ResNet-152 | Acc= 79.50 |
| [19] | Myocarditis | Z-Alizadeh Sani myocarditis dataset | Standard Preprocessing | CNN-KCN | Acc= 97.41 |
| [20] | Myocarditis | Z-Alizadeh Sani myocarditis dataset | Standard Preprocessing | RLMD-PA | Different Results |
| [31] | Myocarditis | Z-Alizadeh Sani myocarditis dataset | Filtering, Image Resize, Cycle GAN | EfficientNet V2 | Acc= 99.33 |
| Proposed Method | Myocarditis | Z-Alizadeh Sani myocarditis dataset | Filtering, Image Resize, New DA method | TNT transformer Model | Acc = 99.71 |

Table (8) shows that the proposed method achieved superior results compared to related works. According to Table (8), it can be seen that limited research has been conducted in the field of DL-based myocarditis diagnosis from CMR images. To demonstrate the efficacy of our proposed method in this study, its results have been compared with those of other papers that focused on CVDs diagnosis. The findings illustrate that our proposed method achieved the highest accuracy level compared to the works reported in the field of CVD diagnosis. Therefore, for future work, our proposed work method can serve as a CADS application for specialist physicians in hospital centers and specialized heart clinics to diagnose myocarditis. For instance, specialist physicians may find it challenging to diagnose various types of CVDs, including myocarditis, in the early stages from CMR images. Therefore, our proposed method can leverage XAI technique to visualize areas that are suspected of myocarditis, which is of great significance for physicians. Moreover, as outlined earlier, our proposed method can be employed to diagnose other types of CVDs. Hence, future research can explore the possibility of using the proposed approach to diagnose all forms of CVDs and achieve successful outcomes. This work can assist specialist physicians in diagnosing a wide range of CVDs with fewer challenges, in the form of a diagnostic software.

Of note, adding some advanced DL architectures for the proposed CADS in this paper is vital. In future, we intend to validate our model with a huge CMRI dataset obtained from more subjects for MCD diagnosis. As another future direction, the provision of MCD datasets comprising a significant number of subjects and diverse classes holds promise for researchers so that they can conduct extensive research in helping to quickly diagnose this disease. In references [97-100], research on CVDs diagnosis using CMRI data and AI techniques are presented. Notably, conventional DL techniques have been widely utilized by researchers to diagnose these conditions. Hence, to achieve successful outcomes in the precise diagnosis of MCD from CMR images, future research could investigate the use of state-of-the-art DL techniques, such as graph CNNs [77-78], compact size CNNs [79], and mutual learning [121].

**Acknowledgement:**
This research is part of the PID2022-137451OB-I00 project funded by the CIN/AEI/10.13039/501100011033 and by FSE+.